\documentclass[acmsmall]{acmart}

\usepackage{amsmath,amsfonts,bm}

\def\eqref#1{equation~\ref{#1}}

\def\1{\bm{1}}

\def\vs{{\bm{s}}}

\DeclareMathAlphabet{\mathsfit}{\encodingdefault}{\sfdefault}{m}{sl}
\SetMathAlphabet{\mathsfit}{bold}{\encodingdefault}{\sfdefault}{bx}{n}

\usepackage{color,xcolor}
\usepackage{epsfig}
\usepackage{graphicx}

\usepackage{adjustbox}
\usepackage{array}
\usepackage{booktabs}
\usepackage{colortbl}
\usepackage{multirow}
\usepackage{subcaption} 
\usepackage{wrapfig}
\usepackage{floatflt}

\usepackage{amsmath,amsfonts}
\usepackage{mathtools}  
\usepackage{bm}
\usepackage{nicefrac}
\usepackage{microtype}

\usepackage{changepage}
\usepackage{setspace}
\usepackage{soul}
\usepackage{xspace}

\usepackage{url}

\usepackage{enumerate}
\usepackage{enumitem}  

\usepackage{makecell}

\usepackage{pifont} 

\usepackage{algorithm,algpseudocode}

\usepackage{amsthm}
\usepackage{framed}
\definecolor{shadecolor}{gray}{0.95}

\usepackage{tcolorbox}

\newcolumntype{L}[1]{>{\raggedright\let\newline\\\arraybackslash\hspace{0pt}}m{#1}}
\newcolumntype{C}[1]{>{\centering\let\newline\\\arraybackslash\hspace{0pt}}m{#1}}
\newcolumntype{R}[1]{>{\raggedleft\let\newline\\\arraybackslash\hspace{0pt}}m{#1}}

\newcommand{\sect}[1]{Section~\ref{#1}}

\newcommand{\fig}[1]{Fig.~\ref{#1}}

\newcommand{\ignore}[1]{}

\makeatletter
\DeclareRobustCommand\onedot{\futurelet\@let@token\@onedot}
\def\@onedot{\ifx\@let@token.\else.\null\fi\xspace}

\def\eg{e.g\onedot} 
\def\ie{i.e\onedot} 
 
\def\etc{etc\onedot}
\def\vs{\emph{vs}\onedot}
\def\wrt{w.r.t\onedot}

\makeatother

\definecolor{MyDarkBlue}{rgb}{0,0.08,1}
\definecolor{MyDarkGreen}{rgb}{0.02,0.6,0.02}
\definecolor{MyDarkRed}{rgb}{0.8,0.02,0.02}
\definecolor{MyDarkOrange}{rgb}{0.40,0.2,0.02}
\definecolor{MyPurple}{RGB}{111,0,255}
\definecolor{MyRed}{rgb}{1.0,0.0,0.0}
\definecolor{MyGold}{rgb}{0.75,0.6,0.12}
\definecolor{MyDarkgray}{rgb}{0.66, 0.66, 0.66}

\newcommand{\model}{SpaLoc\xspace}
\newcommand{\modelfull}{TTTBBBDDD\xspace}

\newcommand{\newtext}[1]{{#1}}

\AtBeginDocument{%
  }

\setcopyright{acmcopyright}
\copyrightyear{2025}
\acmYear{2025}

\acmJournal{CACM}
\acmVolume{1}
\acmNumber{1}
\acmArticle{Communications of The ACM}
\acmMonth{1}

\begin{document}

\title{Building Intelligent Agents with Neuro-Symbolic Concepts}

\author{Jiayuan Mao}
\email{jiayuanm@engineering.upenn.edu}
\affiliation{%
  \institution{University of Pennsylvania}
  \streetaddress{3451 Walnut Street}
  \city{Philadelphia}
  \state{Pennsylvania}
  \country{USA}
}

\author{Joshua B. Tenenbaum}
\email{jbt@mit.edu}
\affiliation{%
  \institution{Massachusetts Institute of Technology}
  \streetaddress{43 Vassar Street}
  \city{Cambridge}
  \state{Massachusetts}
  \country{USA}
}

\author{Jiajun Wu}
\email{jiajunwu@cs.stanford.edu}
\affiliation{%
  \institution{Stanford University}
  \streetaddress{353 Jane Stanford Way}
  \city{Stanford}
  \state{California}
  \country{USA}
}

\renewcommand{\shortauthors}{Mao, Tenenbaum, and Wu}
\renewcommand{\newtext}[1]{#1}

\begin{abstract}
This article presents a concept-centric paradigm for building agents that can learn continually and reason flexibly. The concept-centric agent utilizes a vocabulary of {\it neuro-symbolic concepts}. These concepts, such as object, relation, and action concepts, are grounded on sensory inputs and actuation outputs. They are also compositional, allowing for the creation of novel concepts through their structural combination. To facilitate learning and reasoning, the concepts are typed and represented using a combination of symbolic programs and neural network representations. Leveraging such neuro-symbolic concepts, the agent can efficiently learn and recombine them to solve various tasks across different domains, ranging from 2D images, videos, 3D scenes, and robotic manipulation tasks. This concept-centric framework offers several advantages, including data efficiency, compositional generalization, continual learning, and zero-shot transfer.
\end{abstract}

%

\begin{teaserfigure}
  \includegraphics[width=\linewidth]{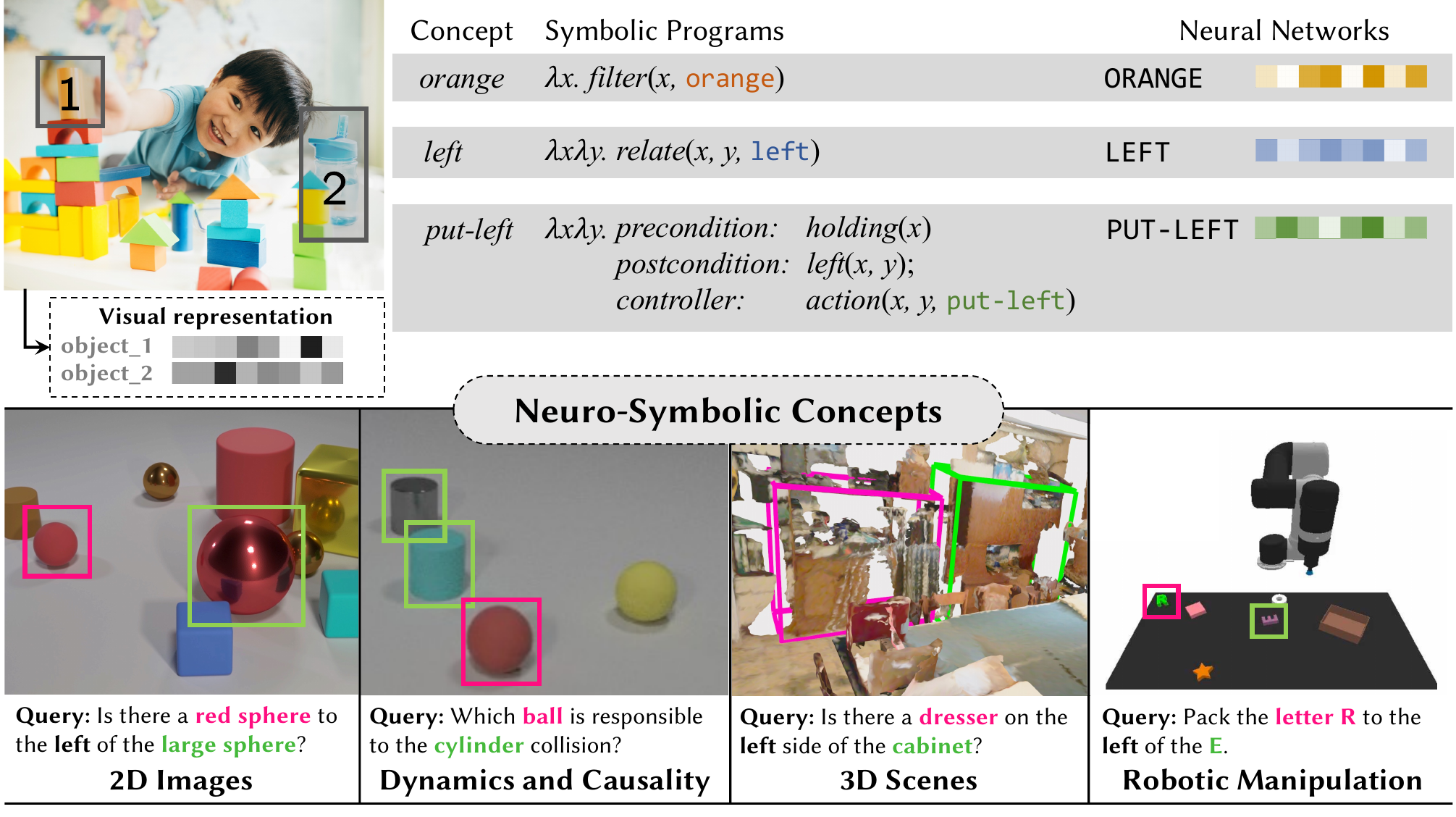}
  \vspace{-20pt}
  \caption{Our framework for building intelligent agents by internalizing a vocabulary of ``concepts,'' which are represented as compositional programs and neural network embeddings. These concepts can be grounded in various domains: 2D images, videos, 3D scenes, and robotic actions, and be recombined based on different types of user queries: visually grounded questions, physical and causal reasoning questions, referring expressions, and manipulation instructions.}
  \label{fig:teaser}
  \vspace{5pt}
\end{teaserfigure}

\received{20 February 2007}
\received[revised]{12 March 2009}
\received[accepted]{5 June 2009}

\maketitle

\begin{tcolorbox}[width=\linewidth, sharp corners=all, colback=white!95!black]
\section*{Key Insights}
\begin{itemize}[leftmargin=*]
\item This article presents a concept-centric paradigm for building agents that can continually learn and reason flexibly.
\item The agent acquires a vocabulary of neuro-symbolic concepts for objects, relations, and actions, represented through a combination of symbolic programs and neural networks. These concepts are grounded in sensory inputs and actuation outputs and can be composed to solve novel tasks using general-purpose reasoning and planning algorithms.
\item The proposed framework offers several advantages, including data efficiency, compositional generalization, continual learning, and zero-shot transfer---key properties for general-purpose AI. These advantages have been demonstrated across applications in vision, language, and robotics.
\end{itemize}
\end{tcolorbox}

\section{Overview}
One of the long-term goals of artificial intelligence (AI) is to build machines that can continually learn new knowledge from their experiences, ground them in the physical world, and apply the knowledge to their reasoning across different tasks, modalities, and environments. The desired capability of such agents includes, but is not limited to, describing perceived scenes, answering queries about scenes, making plans to achieve certain goals, and executing plans in the physical world.
We want such machines to be able to learn and solve a wide variety of tasks across different environments, leveraging a feasible amount of data from multiple modalities.

In recent years, we have seen great success in neural network-based ``end-to-end'' learning methods, but most of them are tailored to particular tasks and environments, for example, categorizing images of objects into a fixed set of labels, translating between particular languages, and playing video games and board games. These systems are usually built on top of relatively simple and monolithic training and inference algorithms (\eg, a single neural network trained by stochastic gradient descent). As a result, their success in domain-specific applications relies on the availability of large-scale datasets and computation resources on the particular task of interest. However, annotating high-quality data for reasoning, planning, and control in visual and physical domains is usually labour-intensive and, in some cases, infeasibly costly. Thus, there has been limited success in extending these methods to building embodied generalist agents across domains. 

The thesis of this article is to emphasize the role of {\it concepts} in learning and reasoning. We draw on rich traditions from philosophy and cognitive science that identify concepts as the basic building blocks of thought~\newtext{(for example, see readings from \citet{margolis1999concepts})}. Humans acquire concepts from the interaction of our evolved cognitive architecture and built-in inductive biases with our own experiences in the world, including both our direct percepts and what we learn socially and culturally from interacting and communicating with other humans. Our minds then compose these basic units to form sophisticated compound thoughts: beliefs, desires, and plans. One of the most powerful ways to construct a system of useful concepts for reasoning is to build them out of meanings acquired through language. In particular, we can consider the granularity of concepts at the level of individual word meanings (meanings of nouns, verbs, \etc), and treat their combinations in a similar way as how words can be combined into phrases and sentences in natural language. \newtext{Similar ideas of composing primitive units of thought have a long tradition in AI, dating back to inductive logic programming~\cite{muggleton1994inductive}, statistical relational learning~\cite{friedman1999learning}, and probabilistic-logic programming~\cite{manhaeve2018deepproblog}}.

%
Our technical proposal in this paper is to build a {\it neuro-symbolic concept} representation. Each concept is a discrete symbol (word or short phrase) that can be grounded onto subsets of the embodied environments. An object concept ``orange'' grounds to the sets of orange objects; an object relation concept ``left'' grounds to all object pairs (A, B) such that A is left of B; an action concept ``put-left'' grounds to all agent action sequences that move the object currently being held to a position on the left of the reference object. This includes pushing the object to the goal, holding it vigorously using tools, and so forth. As illustrated in \fig{fig:teaser}, during reasoning, our {\it concept-centric} framework operates at a higher level of abstraction with this vocabulary of ``{\it neuro-symbolic concepts}.''

Neuro-symbolic concepts have two advantages: they can be flexibly grounded on sensory and actuation modalities, and they have strong compositional generalization, since existing concepts can be structurally combined to form new ones. For instance, ``orange,'' ``cylinder,'' and ``left'' can be combined to form a novel concept such as ``put the orange cylinder to the left of the bottle.'' Such compositionality naturally suggests a decomposition of the learning problem: we may individually learn how to recognize object shapes (cylinder, bottle), how to recognize object colors (orange), how to reason about object placement (left of the target), and how to move objects from one location to another. Finally, during inference time, we can recombine these learned concepts: shapes, colors, relations, and actions, to achieve the specified goals.

Compared with purely end-to-end learning methods, the inherent compositionality of neuro-symbolic concepts makes them much better suited for generalist agent learning. Compositionality simultaneously supports four properties required by generalist agents: data efficiency (because collecting labeled data is especially difficult), compositional generalization (the number of possible scenes and tasks can be exponentially large as the number of objects and their properties increases; we want systems that can generalize to unseen scenes and unseen goals), continual learning (the system should be able to learn and adapt gradually), and transfer learning (we want to support transfer among different tasks). 

In the rest of the article, we will first offer a definition of neuro-symbolic concepts (\sect{sec:concepts}). We will then delve into a concrete framework of neuro-symbolic concept learning for visual scene understanding and showcase its ability to learn from few data, continually and generalizably, and be transferable (\sect{sec:learning}). Finally, we will discuss other applications of the approach (\sect{sec:applications}).

\section{Neuro-Symbolic Concepts}
\label{sec:concepts}

\begin{figure}[tp]
    \centering
    \includegraphics[width=\linewidth]{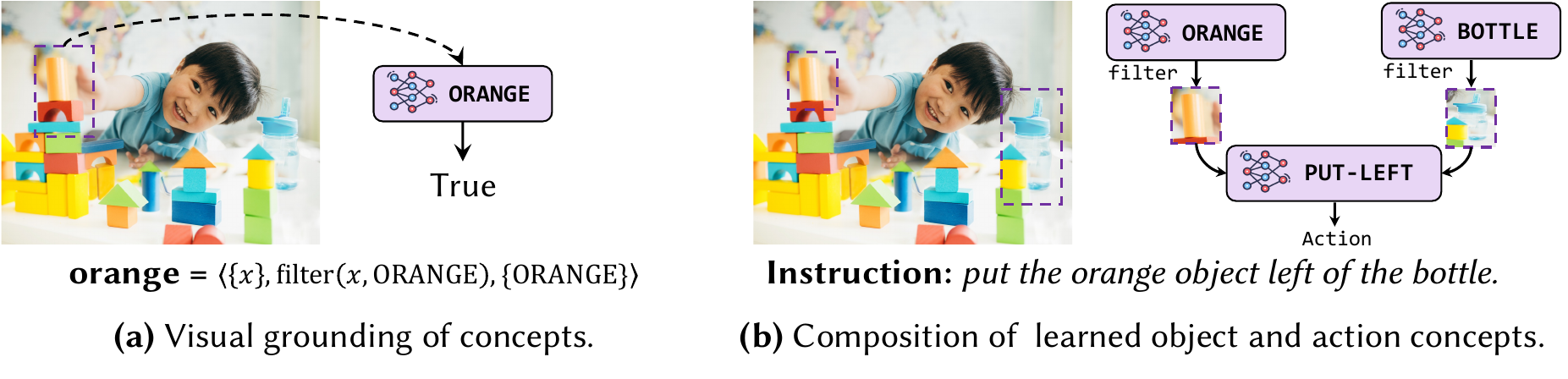}
    \caption{In a neuro-symbolic concept-centric framework, different concepts such as object categories, properties, relations, and actions are represented as a combination of programmatic and neural representations. In (a), the neural representation connects the concept with sensory and actuation representations. In (b), different concepts can be combined to form new compound concepts.
}
    \label{fig:teaser-comparison}
\end{figure}

In order to systematically represent the grounding of concepts and how they can be composed, we formally represent each concept $c$ as a tuple of 
\[
    c = \langle \textit{parameter}, \textit{program}, \textit{neural-nets}\rangle .
\]
In essence, our concept representation integrates neural network representations (for instance, vector embeddings), which ground concepts in visual and physical representations, and symbolic representations (particularly, parameterized programs), which characterize how various concepts can be combined. Shown in \fig{fig:teaser-comparison}a, the concept ``orange'' is represented as 
\[
\textit{orange} = \langle \{x\}, \textit{filter}(x, \textbf{ORANGE}), \{ \textbf{ORANGE} \} \rangle,
\]
where $\textbf{ORANGE}$ denotes a vector embedding that can be used by the built-in \textit{filter} function to classify orange objects. In practice, this $\textit{filter}$ function can be implemented by computing the cosine similarity between the neural representation of object $x$ and $\textbf{ORANGE}$~\citep{Mao2019NeuroSymbolic}. 

Similarly, a relational concept (\eg, a prepositional phrase) ``left-of'' can be defined as 
\[\textit{left-of} = \langle \{x, y\}, \textit{relate}(x, y, \textbf{LEFT-OF}), \{ \textbf{LEFT-OF} \} \rangle.\]
The concept relates two objects $x$ and $y$. The $\textit{relate}$ function will compute the cosine similarity between $\textbf{LEFT-OF}$ and a pairwise representation between $(x, y)$ to determine whether $x$ is left of $y$ in the scene. Actions (\eg, verbs) would have three parts in their programs: a controller that can generate sequences of robot control commands, and the pre- and post-conditions for the action. For example, an action ``put-left-of'' can be represented as:
\begin{align*}
    \textit{put-left-of} = \langle &\{x, y\}, \{ \textit{pre}=\textit{holding}(x), \textit{post}=\textit{left-of}(x, y), \\
    &\textit{controller}=\textbf{PUT-LEFT-OF} \}, \{ \textbf{PUT-LEFT-OF} \} \rangle.
\end{align*}
Here, the preconditions and post-conditions of the action can be described with formulas composed from other object-level and relational concepts.

Illustrated in \fig{fig:teaser-comparison}b, this representation of concepts enables us to combine existing concepts adhering to symbolic functional composition rules, to form compound concepts such as $\textit{orange}(x)~\textbf{and}$ $\textit{cylinder}(x)$ (orange cylinders), or more complex ones such as $\textit{orange}(x)$ $\textbf{and}$ $\textit{bottle}(y)$ $\textbf{and}$ $\textit{put-left-}$\\$\textit{of}(x, y)$ (put the orange object left of the bottle). To support the formal compositionality of different concepts, all parameters and outputs are typed with primitive types (including objects, events, actions, Booleans, and integers). For instance, object concepts are represented as functions that take the perceptual representation of an object as input and predict classification scores as output, indicating whether the object has the concept. Relational concepts are associated with classifiers that classify object pairs. Meanwhile, action concepts are linked with preconditions (circumstances under which the action can be executed), postconditions (the outcomes of executing the action), and controllers that generate agent actions based on the current perceptual state. Depending on the domain and the task, one can choose to implement different primitive operations (\eg, \textit{filter} in a visual recognition context).


\begin{figure*}[tp]
    \centering
    \includegraphics[width=\linewidth]{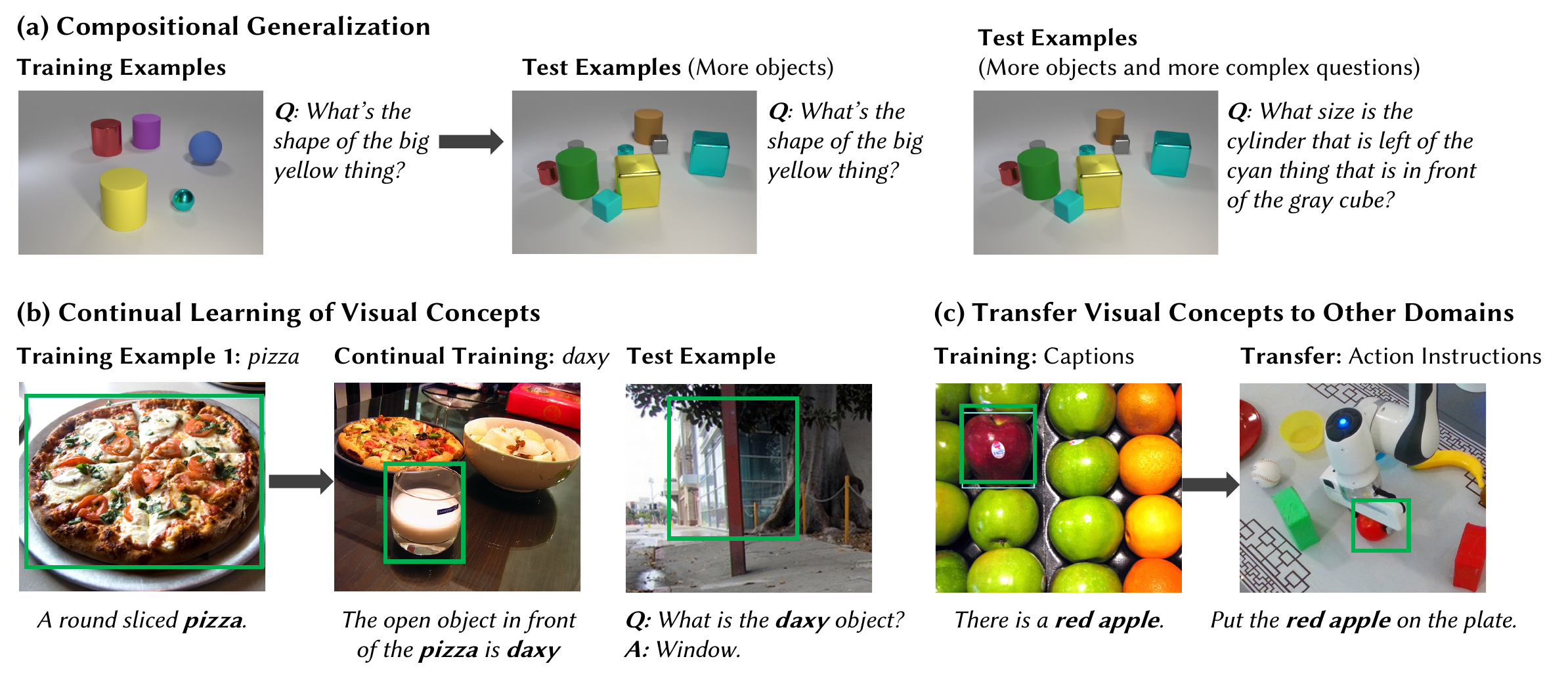}
    \caption{Three challenges in (a): Compositional generalization. (b): Continual learning of concepts for reasoning. (c): Transfer learned concepts across domains.
    }
    \label{fig:teaser-challenges}
\end{figure*}

Leveraging such neuro-symbolic concepts, we can efficiently and effectively learn the grounding of concepts in various domains and recombine them to solve different downstream tasks. For example, shown in \fig{fig:teaser}, neuro-symbolic concepts can be grounded in 2D images (such as object properties and object relations), videos (physical events and their relations), 3D scenes (object properties and viewpoint-dependent relations), and finally, robotic manipulation tasks (object properties, relations, and actions that change them). Therefore, by recombining them through symbolic program structures, we can answer questions, resolve referring expressions, and interpret human instructions. We illustrate this idea in \fig{fig:teaser-comparison}. Specifically, we learn the grounding (classifiers and controllers) for individual visual and action concepts and recombine them following a hierarchical program that represents the meaning of the input user query: put the orange object left of the bottle. In the following, we revisit the four important desiderata for generalist agent learning and illustrate how our neuro-symbolic concept-centric paradigm fulfills all requirements.

\paragraph{Data efficiency.} Since learning in embodied environments inevitably involves machine interaction with the physical environment and human annotations, minimizing the amount of data needed to learn a specific concept is crucial. Compared to a monolithic deep neural network, a concept-centric framework gains data efficiency primarily by leveraging modular structures of the learning task. For example, the complex concept ``push the orange cylinder'' can be decomposed into three individual concepts. Such decomposition structure injects strong prior to the learning algorithm that the whole concept is the conjunction of two object concepts (``orange'' and ``cylinder'') and an action concept (``push''). It further enables the algorithm to disentangle the learning problem and perform explicit multi-task learning from various sources (\eg, learning the concept of ``orange'' from images and ``push'' from robot learning datasets).

\paragraph{Compositional generalization.}
Compositionality is often grounded in different aspects across different domains. For example, in the visual concept learning domain, as illustrated in \fig{fig:teaser-challenges}a, compositional generalization is at least expected at two levels: the concept composition level (\eg, ``the big cylinder left of the yellow block''), and the scene composition level (generalization to scenes with a different number of objects compared to training examples).
The advantage of concept-centric frameworks is primarily a contribution from the alignment between the composition structure of concepts and the structure of the domain. For example, explicitly reasoning about the set of ``orange'' and its subset ``orange cylinder'' in a visual scene makes the reasoning process robust to the total number of objects in a scene.

\paragraph{Continual learning.}
The demand for continual concept learning and transfer learning arises due to two challenges. First, it is generally difficult to obtain high-quality ``end-to-end'' learning examples for embodied agents: \ie, from raw perceptual input to robot control commands. Therefore, a practical system should be able to learn from multiple sources of data, typically of different input-output specifications: unannotated videos, paired images and texts, human demonstrations of skills, and so on. Second, at deployment time, the machine should continually adapt itself to its environment, such as learning new concepts (\eg, an unseen breed of dogs, a new type of dish, \etc) and new human preferences.

\paragraph{Zero-shot transfer.}
Unlike monolithic deep neural networks that usually require tuning all parameters while learning new concepts, as illustrated in \fig{fig:teaser-challenges}b, the concept-centric framework naturally allows a flexible introduction of new concepts and adjustment to a single previously learned concept, thanks to the modular structures of concept composition. It also enables a zero-shot transfer of learned concepts across tasks and even domains, such as transferring learned object concepts from the task of image captioning (``the photo shows a dog'') to visual question answering (``how many dogs are there?''), from the domain of visual concept learning (``apples'') to the domain of robotic manipulation (``push the apples''), as illustrated in \fig{fig:teaser-challenges}c.


\section{Learning Neuro-Symbolic Visual Concepts}
\label{sec:learning}

\renewcommand{\modelfull}{Neuro-Symbolic Concept Learner\xspace}
\renewcommand{\model}{NS-CL\xspace}

\begin{figure*}[tp]
    \centering
    \includegraphics[width=\linewidth]{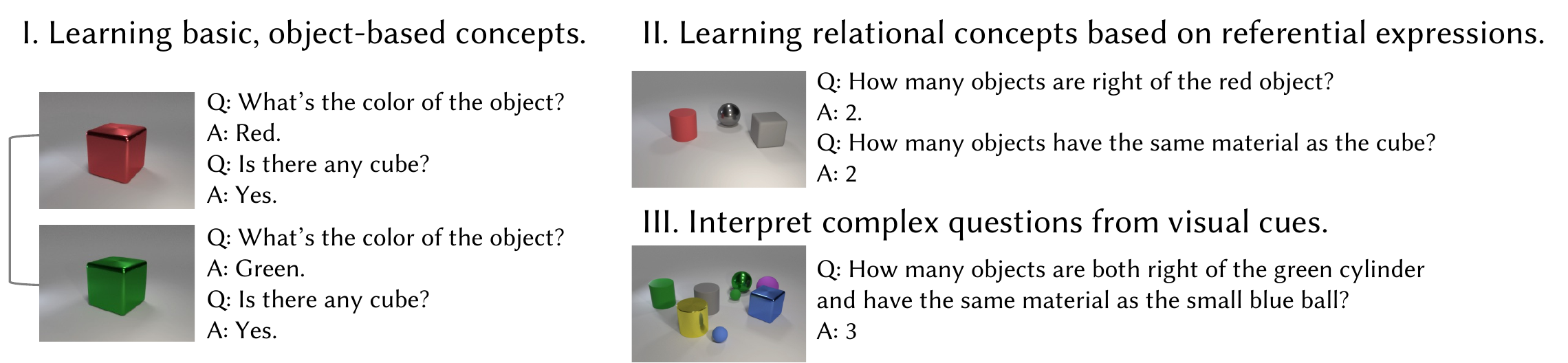}
    \caption{Humans learn visual concepts, words, and semantic parsing jointly and incrementally. {\bf I.} Learning visual concepts (red \vs green) starts from looking at simple scenes, reading simple questions, and reasoning over contrastive examples~\citep{fazly2010probabilistic}. {\bf II.} Afterwards, we can interpret referential expressions based on the learned object-based concepts, and learn relational concepts (\eg, on the right of, the same material as). {\bf III} Finally, we can interpret complex questions from visual cues by exploiting the compositional structure.}
    \label{fig:nscl-teaser}
\end{figure*}

Our framework for neuro-symbolic concept learning for visual scene understanding is motivated by how humans learn visual concepts by jointly understanding vision and language \citep{fazly2010probabilistic}. Consider the example shown in \fig{fig:nscl-teaser}-I. Imagine someone with no prior knowledge of colors is presented with the images of the red and green cubes, paired with the questions and answers. They can easily identify the difference in objects' visual appearance (in this case, color), and align it to the corresponding words in the questions and answers (\textit{Red} and \textit{Green}). Other object attributes (\eg, shape) can be learned in a similar way. Starting from there, humans are able to inductively learn the correspondence between visual concepts and word semantics (\eg, spatial relations and referential expressions, \fig{fig:nscl-teaser}-II), and unravel compositional logic from complex questions assisted by the learned visual concepts (\fig{fig:nscl-teaser}-III, also see \cite{abend2017bootstrapping}). 

This motivated us to build a learning framework that jointly learns visual perception, words, and semantic language parsing from images and question-answer pairs. Proposed in \citet{Mao2019NeuroSymbolic}, a \modelfull (\model) learns all these from natural supervision (\ie, images and QA pairs), requiring no annotations on images or semantic programs for sentences. Instead, analogous to human concept learning, it learns via curriculum learning. \model starts by learning representations/concepts of individual objects from short questions (\eg, What's the color of the cylinder?) on simple scenes ($\leq$3 objects). By doing so, it learns object-based concepts such as colors and shapes. \model then learns relational concepts by leveraging these object-based concepts to interpret object referrals (\eg, Is there a box right of a cylinder?). The model iteratively adapts to more complex scenes and highly compositional questions.

\begin{figure*}[tp]
    \centering
    \includegraphics[width=\linewidth]{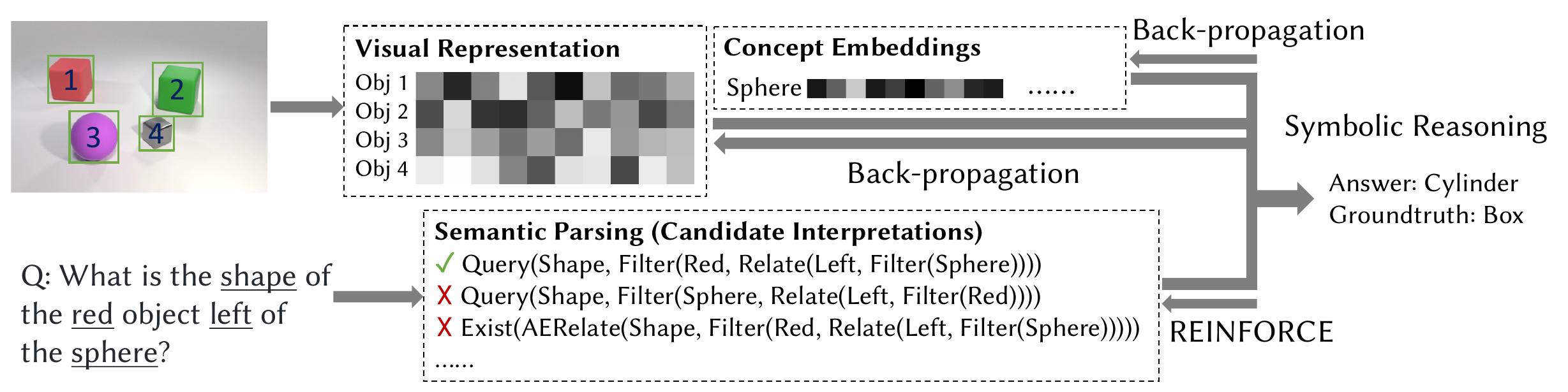}
    \caption{NS-CL uses neural symbolic reasoning to bridge the learning of visual concepts, words, and semantic parsing.}
    \label{fig:nscl}
\end{figure*}

Shown in \fig{fig:nscl}, \model has three modules: a neural network-based perception module, a semantic parser for translating questions into executable programs, and a symbolic program executor. Given an input image, the visual perception module detects objects in the scene and extracts a deep, latent representation for each of them. The semantic parsing module translates an input question in natural language into an executable program,  represented in a domain-specific language (DSL) designed for VQA. The DSL covers a set of fundamental operations for visual reasoning, such as filtering out objects with certain concepts or querying the attribute of an object. The generated programs have a hierarchical structure of symbolic, functional modules.


\begin{figure}[tp]
    \centering
    \includegraphics[width=\linewidth]{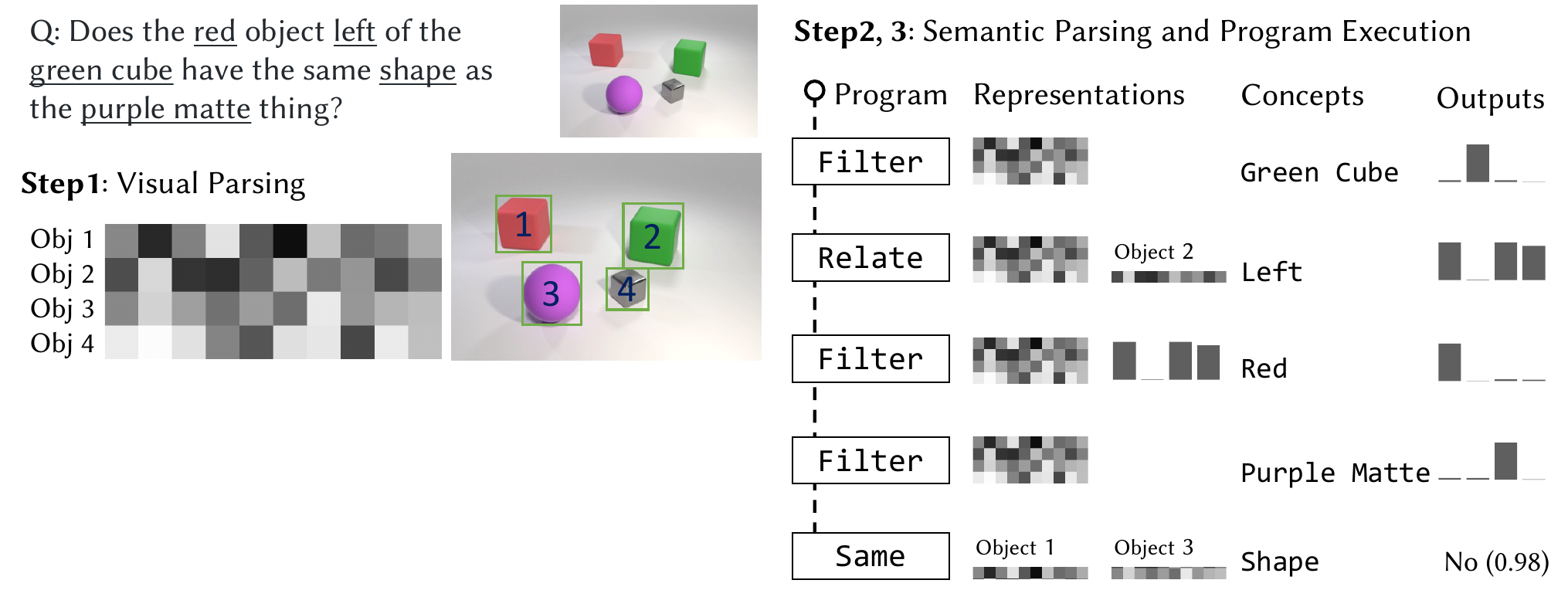}
    \caption{The neuro-symbolic execution procedure of a program based on the visual representation and concept embeddings.}
    \label{fig:nscl-execution}
\end{figure}

Next, based on the latent program recovered from the question in natural language, a symbolic program executor executes the program and derives the answer based on the object-based visual representation. Our program executor mainly contains two parts: the concept quantization module and a collection of deterministic functional modules. The concept quantization module classifies object attributes and relations, and the functional modules implement the logic of composing these classification results.
To make the execution differentiable \wrt visual representations, we represent the intermediate results in a probabilistic manner: a set of objects is represented by a vector, as the attention mask over all objects in the scene. Each element, $\mathrm{Mask}_i \in [0, 1]$ denotes the probability that the $i$-th object of the scene belongs to the set. \fig{fig:nscl-execution} shows an illustrative execution trace of a program. The first \textit{filter} operation outputs a mask of length $4$ (there are in total four objects in the scene), with each element representing the probability that the corresponding object is selected (\ie, the probability that each object is a green cube). The output ``mask'' on the objects will be fed into the next module (\textit{relate} in this case) as input and the execution of programs continues. The last module outputs the final answer.

\begin{figure*}[tp]
    \centering
    \includegraphics[width=\linewidth]{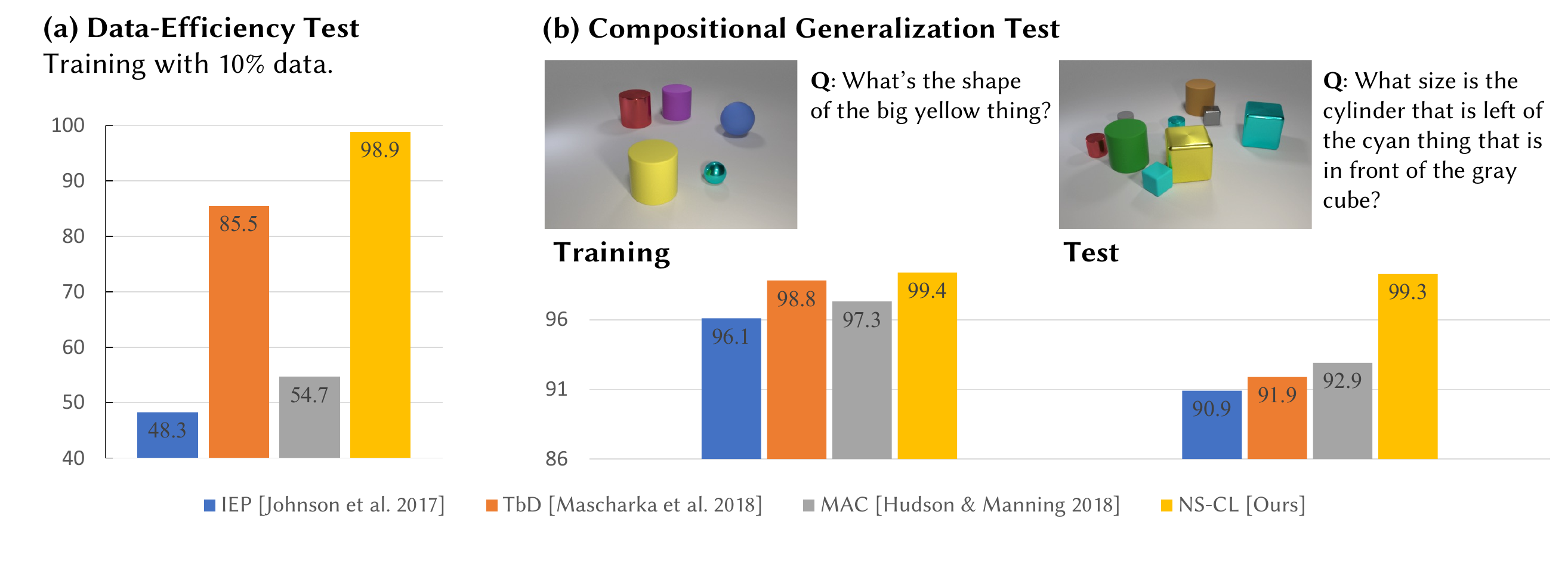}
    \caption{Data efficiency test and compositional generalization test for the \modelfull (\model).}
    \label{fig:nscl-result}
\end{figure*}

\paragraph{Data efficiency.} \model's modularized design enables interpretable, robust, and accurate visual reasoning: shown in \fig{fig:nscl-result}a, it achieves state-of-the-art performance on the CLEVR dataset~\citep{Johnson2017CLEVR}. More importantly, it enables data-efficient learning of concepts and combinatorial generalization \wrt both visual scenes and semantic programs. Highlighted in \fig{fig:nscl-result}, compared to other approaches that do not explicitly learn concepts, (a) when trained on 10\% of the CLEVR training data, it achieves 98.9\% accuracy on the test set, surpassing all baselines by 14\%.

\paragraph{Compositional generalization.} We also test our model for compositional generalization; shown in \fig{fig:nscl-result}b, after being trained on scenes with a small number of objects and simple questions, \model directly generalizes to more complex scenes and questions, while all baselines show a significant performance drop.

\paragraph{Continual learning.} Since our neuro-symbolic learning problem decomposes the learning problem into learning individual concepts, it naturally supports continual learning of new concepts. As a concrete implementation, in \citet{Mei2022Falcon}, we present a meta-learning framework for learning new visual concepts quickly from just one or a few examples, guided by multiple naturally occurring data streams: simultaneously looking at images, reading sentences that describe the objects in the scene, and interpreting supplemental sentences that relate the novel concept to other concepts. \newtext{The system operates in a class-incremental manner~\citep{van2022three}, where it continuously receives new examples of an unseen category and builds a new embedding for the novel category.} The learned concepts support downstream applications, such as answering questions by reasoning about unseen images. Our model, namely FALCON, represents individual visual concepts, such as colors and shapes, as embeddings in a high-dimensional space. Given an input image and its paired sentence, our model first resolves the referential expression in the sentence and associates the novel concept with particular objects in the scene. Next, our model interprets supplemental sentences to relate the novel concept with other known concepts, such as ``X has property Y'' or ``X is a kind of Y''~\citep{Han2019Visual}. Finally, it infers an optimal embedding for the novel concept that jointly 1) maximizes the likelihood of the observed instances in the image, and 2) satisfies the relationships between the novel concepts and the known ones. We demonstrate the effectiveness of our model on both synthetic and real-world datasets.

\paragraph{Transfer learning.} The learned visual concepts can also be used in other domains, such as image retrieval. With the visual scenes fixed, the learned visual concepts can be transferred directly into the new domain. We only need to learn the semantic parsing of natural language into the new DSL. We build a synthetic dataset for image retrieval. The dataset contains only simple captions: ``There is an $<$object A$>$ $<$relation$>$ $<$object B$>$.'' (\eg, There is a box right of a cylinder). The semantic parser learns to extract corresponding visual concepts (\eg, \texttt{box}, \texttt{right}, and \texttt{cylinder}) from the sentence. The program can then be executed on the visual representation to determine if the visual scene contains such relational triples. Note that this functionality cannot be directly implemented on the CLEVR VQA program domain, because questions such as ``Is there a box right of a cylinder'' can be ambiguous if there exist multiple cylinders in the scene. Due to the entanglement of the visual representation with the specific DSL, baselines trained on CLEVR QA cannot be applied directly to this task. 

\section{Applications}
\label{sec:applications}

The design principles of \model, in particular, the visual grounding of concepts through neuro-symbolic reasoning, can naturally generalize to a large body of learning and reasoning tasks. 

\paragraph{Accurate and robust image captioning.} \citet{Wu2019Unified} implement a similar idea of neuro-symbolic concept learning to image-caption retrieval tasks. They used pretrained language parsers to translate captions into graphical representations composed of object categories, properties, and relationships. This kind of factorization not only leads to better performance in retrieving accurate descriptions of images, but also improves the robustness of the system with respect to captions that are similar to correct ones (\eg, differ only in one or two words) but inaccurate.

\paragraph{Video and counterfactual reasoning.} A line of research~\citep{Chen2021DCL,ding2021dynamic} has been extending the concept learning framework to reasoning about physics. The object-centric nature of neuro-symbolic concept learners enables natural integration with learned physics models, which brings the capability to perform predictive and counterfactual reasoning. As an example, in \citet{Chen2021DCL}, the authors ground concepts about physical objects and events from dynamic scenes and language. Building upon a neural object-centric representation, their model is simultaneously also trained to approximate the dynamic interaction among objects with neural networks. Therefore, after training, it can not only detect and associate objects and events across the frames, but also make future and counterfactual predictions of object interactions (\eg, ``what will happen if we remove the red block from the scene?''). Later work such as \citet{ding2021dynamic} further extends this capability to online inference of object physical properties.


\paragraph{3D concept grounding.} The neuro-symbolic framework can also be applied to 3D representations~\citep{prabhudesai2020disentangling,Hsu2023NS3D}. The variability of the 3D domain induces two fundamental challenges: 1) the expense of labeling and 2) the complexity of 3D grounded language. Hence, essential desiderata for models are to be data-efficient, generalize to different data distributions and tasks with unseen semantic forms, as well as ground complex language semantics (\eg, view-point anchoring and multi-object reference: ``facing the chair, point to the lamp on its right''). To address these challenges, \citet{prabhudesai2020disentangling} propose to disentangle different object properties in learning to achieve better performance in few-shot concept learning. \citet{Hsu2023NS3D} extend the neuro-symbolic concept learner model by introducing functional modules that effectively reason about high-arity relations (\ie, relations among more than two objects), key in disambiguating objects in complex 3D scenes. This architecture enables significantly improved performance on settings of data efficiency and generalization, and demonstrates zero-shot transfer to a 3D question-answering task. 

\paragraph{Human motion.} \citet{endo2023motion} focuses on designing models that conduct complex spatiotemporal reasoning over motion sequences. \citeauthor{endo2023motion} proposes a new framework for learning neural concepts of motion, attribute neural operators, and temporal relations. Unlike 2D and 3D vision domains, where object segmentations can be readily extracted using pre-existing object detectors, motion sequences lack a universal action segmentation methodology. Therefore, they propose to jointly temporally localize and ground motion concepts.

\paragraph{Robotic manipulation.} Recall that an important feature of neuro-symbolic concept learning methods is that the learnable modules associated with different concepts are naturally disentangled. Therefore, it directly supports the transfer of learned concepts to other tasks or even domains (\eg, from vision-language domains to robotic manipulation domains). In \citet{Ren2023ProgramPort} and \citet{namasivayam2023learning}, the authors tackle the problem of learning robotic manipulation based on visual input. Both papers exploit the syntactic and semantic structures of language instructions to build robotic manipulation algorithms composed of object recognition models and action policies. \citet{namasivayam2023learning} directly transfers the visual concepts learned by the neuro-symbolic concept learning on images to robotic manipulation, by learning additional object movement policies with reinforcement learning. \citet{Ren2023ProgramPort}, by contrast, leverages large-scale pretrained vision-language (VL) models for object property recognition. Compared to a conventional pretraining-finetuning pipeline for leveraging pretrained models for robotics, their method leads to more data-efficient learning and, more importantly, better zero-shot generalization in a variety of unseen objects and tasks.



\section{General Discussion}
We have presented a general framework for learning and reasoning that is applicable to various domains and tasks. By leveraging the neuro-symbolic concept representation, our system can continuously learn concepts from data streams in a data-efficient manner and programmatically compose its learned representations to solve new tasks, even previously unseen tasks. This capability allows us to learn and generalize concepts from diverse types of data streams, including image-caption data and robotic demonstrations, in order to solve complex tasks.

\newtext{
Our neuro-symbolic concept learning framework belongs to the broader paradigm of neuro-symbolic AI, a field where researchers explore the synergies between neural networks, symbolic reasoning methods, and probabilistic inference tools. The idea of connecting neural networks with symbolic entities has its origins in early work on embedding symbolic relationships into vector representations. For example, early research~\citep{rocktaschel2015injecting,demeester2016lifted} demonstrated how to integrate logical reasoning and neural networks, improving the efficiency, interpretability, and controllability of learning systems.
}

\newtext{
Building on top of these high-level ideas, and in line with NS-CL, many neuro-symbolic AI systems have been developed that combine symbolic reasoning mechanisms with neural networks for recognizing object properties and relationships, as well as predicting action commands in interactive environments. By combining perceptual capabilities with tools like forward-chaining theorem provers, answer set programming solvers, and program synthesis tools, these frameworks enable reasoning and planning in both visual and physical environments.
}

\newtext{
In application areas closely related to NS-CL, \citet{dfol_vqa_icml2020} introduced differentiable first-order logic frameworks for reasoning about objects in scenes.
Further, \citet{barbiero2023interpretable} combined neural network predictions with fuzzy-logic rule execution, and \citet{shindo2023alpha} used differentiable inductive logic programming to recover logic programs representing scene structures for scene reasoning tasks. These methods further formalize the interface between neural and symbolic components using tools like probabilistic and real-valued logic, supporting formal interpretations of reasoning under uncertainty. These approaches excel not only in data efficiency and compositional generalization but also in offering interpretability of reasoning traces and inferred rules, which is critical in applications that demand transparency.
}

\newtext{
While much of the prior work has focused on learning simple rules or answering queries about object states and relationships, recent research has extended neuro-symbolic frameworks to more complex reasoning tasks. For example, \citet{wang2019satnet} and \citet{yang2023neurasp} tackled visual puzzle-solving, such as Sudoku, using Boolean satisfiability solvers and answer set programming tools. End-to-end neural networks for these complex tasks often require significantly more data, which can be difficult to obtain for many practical applications. However, training such complex neuro-symbolic can be challenging, as backpropagation must occur over long chains of neuro-symbolic computations without intermediate supervision.
}

\newtext{
Moving toward more abstract, layout-based, and scene-level concepts --- an emerging area in machine learning and visual reasoning--- \citet{shindo2023learning} have studied how patterns of object placements can be learned from just a few examples.
Similarly, \citet{hsu2024makes} investigated reasoning about abstract concepts like mazes and treasure maps.
These high-level concepts are difficult to interpret by state-of-the-art end-to-end systems like large vision-language models. These studies have shown that decomposing abstract concepts into smaller, more primitive entities using symbolic structures can significantly improve system performance.}

\newtext{
Finally, it is important to highlight other key advantages of neuro-symbolic systems that we have not fully discussed in this paper, such as interpretability, controllability, and the ability to integrate with external knowledge bases~\cite{barbiero2023interpretable,skryagin2023scalable}. These attributes, especially interpretability, and safety, are critical in high-stakes decision-making contexts~\cite{yang2023safe}.
}

The idea of neuro-symbolic concepts is also closely related to the idea of neural module network compositions~\citep{andreas2016neural}, since computationally, they are both paradigms for composing neural network modules to solve more complex tasks. However, they differ at both the conceptual and implementation levels. Specifically, in neural module networks, primitive neural networks define functions or transformations that can be applied to inputs, whereas in neuro-symbolic concepts, primitive neural networks handle the grounding of individual concepts, and the operations based on these concepts (\eg, \textit{filter} or \textit{count}) are implemented as deterministic functions in domain-specific languages (DSLs). This disentanglement between grounding and reasoning brings about significant improvements in data efficiency, compositional generalization, and transferability. 

Of course, such improvements come at a cost: many works on neuro-symbolic concept learning have the limitation of relying on a predefined DSL. This DSL encompasses primitive operators such as \textit{filter} and \textit{relate}, as well as concept symbols such as \textit{orange} and \textit{place}. In the following, we are going to delve into these two parts.

In most domains, by combining object property primitives, relational primitives, and action primitives, along with simple set operations such as intersection, union, and counting, we can construct a highly capable system. Recently, there has been a growing interest in extending the primitive set and number operations to general programming languages, such as Python code~\citep{surismenon2023vipergpt}. This will greatly improve the expressiveness of the programs by including complex control flows such as loops and recursions. However, in general, this also introduces new challenges in the learning of concepts. Recall that in NS-CL and many neuro-symbolic concept learning works, the concept representations are learned through back-propagation of the program execution trace. Although there has been much work on backpropagation against program execution traces~\citep{petersen2021learning}, complex control flows will inevitably introduce intractability of possible execution results, as well as gradient vanishing and explosion problems.

Relying on a predefined set of concepts can be too restrictive in many real-world applications. Three approaches have emerged to address this: grammar-based lexicon learning~\citep{Mao2021Grammar} and more recent approaches based on large language models (LLMs). The high-level idea behind grammar-based lexicon learning is that instead of trying to associate each word or phrase with a predefined set of concept names (\ie, mapping the word ``orange'' to a concept ``ORANGE''), we construct the library of concepts by ``converting'' each word into a concept. This is roughly equivalent to discovering the syntax of each word. For example, if we see the word ``orange'' in a sentence and it is an adjective, then we immediately know the word ``orange'' should correspond to an object property concept, named ``ORANGE.'' This approach is called ``grammar-based lexicon learning'' because the system begins with a small set of universal grammar rules and jointly discovers new concepts from the text corpus while learning their grounding.

\newtext{A second approach involves inducing new concepts from experiences. One method is to represent concepts as ``theories composed of other concepts,'' in line with the theory-theory of concepts~\cite{morton1980frames}. For example, \citet{das2020few} introduces new logical concepts by composing previously learned concepts using logic programs. Likewise, \citet{shindo2023learning} induces transferable scene-level concepts from a few examples, while \citet{ellis2023dreamcoder} learns a ``library'' of functions built from primitive concepts that can be hierarchically recombined to form complex geometric and scene-level concepts. Another direction explores the invention of new object and relation concepts through contrastive learning~\cite{sha2024neuro}.}

A third approach is to leverage large language models (LLMs) such as GPT-4. With the success of large language models in language-to-code translation, researchers have also explored the use of these models in extracting concept symbols from natural language queries~\citep{Hsu2023NS3D,surismenon2023vipergpt}. In particular, they leverage large language models that have been trained on Internet-scale text and code corpora to translate natural language queries into programs with concept symbols. These concept symbols are not chosen from a given vocabulary, but are instead automatically generated by LLMs based on the user queries. For each concept symbol that appears in LLM-translated programs, a new concept representation will be initialized and learned.

\newtext{There are many challenges and future directions for neuro-symbolic concept learning systems. Most approaches focus on relational concepts involving only two objects (or in the case of 3D concepts~\citep{Hsu2023NS3D}, three), but there are more complex layout concepts and scene-level concepts (e.g., mazes~\cite{hsu2024makes}) that involve many more objects and have variable arities, raising the question of how to handle such complexity. Additionally, while NS-CL uses curriculum learning, and some work has explored different methods for curriculum construction~\cite{li2020competence}, the automatic discovery or design of curricula that adapt as humans do when learning new concepts remains unsolved. Next, so far, many concept learning systems can only operate in a pure ``class-incremental'' learning framework~\cite{van2022three}, where new concepts build on previous ones, but addressing the full ``curriculum learning'' setting --- requiring revision or reversion of previously learned concepts --- poses another significant challenge. Moreover, how to enable unsupervised concept learning, beyond the current focus on supervised or semi-supervised methods, is an important goal.
Finally, another direction for future work is to formalize reasoning under perceptual and other types of uncertainty by incorporating probabilistic inference methods. This includes tools like probabilistic logic programming and probabilistic programming languages.}

So far, most neuro-symbolic concept learning systems have been developed for particular domains and tasks. For example, we have systems that can learn and reason with concepts for 2D images~\citep{Mao2019NeuroSymbolic}, 3D scenes~\citep{Hsu2023NS3D}, human motions~\citep{endo2023motion}, video events~\citep{Chen2021DCL}, and robotic actions~\citep{Ren2023ProgramPort,mao2022pdsketch}. However, they have been built in isolation, with limited shared knowledge among them. An important future direction for neuro-symbolic concept learning is the development of scalable, cross-domain concept libraries. From an engineering perspective, similar to recent large language models such as OpenAI GPT and Google Gemini, which can solve a wider range of tasks in the language domain based on user instructions, building unified concept representations across domains and modalities would enable us to tackle a broader spectrum of embodied AI problems, spanning from perception to action. From a scientific perspective, connecting and even unifying concept representations across domains and modalities could not only bring better data efficiency in learning, but also enable the grounding of concepts in more abstract scenarios. For example, the concept ``close to'' has grounded meanings across different domains, but the core and abstract notion of distance metrics is really shared across all domains and modalities.

\begin{acks}
This work is in part supported by AFOSR YIP FA9550-23-1-0127, FA9550-22-1-0387, ONR N00014-23-1-2355, ONR YIP N00014-24-1-2117, ONR MURI N00014-22-1-2740, and NSF RI \#2211258, and an AI2050 Senior Fellowship.
\end{acks}

\bibliographystyle{ACM-Reference-Format}
\bibliography{reference}

\section*{Author Information}
Jiayuan Mao is an assistant professor at the University of Pennsylvania, Philadelphia, Pennsylvania, USA. \\
Joshua B. Tenenbaum is a professor at the Massachusetts Institute of Technology, Cambridge, Massachusetts, USA. \\
Jiajun Wu is an assistant professor at Stanford University, Stanford, California, USA.

\end{document}